\documentclass[conference,onecolumn,draft]{IEEEtran}
\IEEEoverridecommandlockouts
\usepackage{cite}
\usepackage{amsmath,amssymb,amsfonts}
\usepackage{graphicx}
\usepackage{textcomp}
\usepackage{xcolor, pifont}

\usepackage{authblk}
\usepackage{multirow}
\usepackage{stfloats}
\usepackage{subcaption}

\usepackage{algcompatible}
\usepackage[ruled,vlined,linesnumbered]{algorithm2e}
\algblockdefx{FORALLP}{ENDFAP}[1]%
  {\textbf{for all }#1 \textbf{do in parallel}}%
  {\textbf{end for}}

\newcommand*\colourcheck[1]{%
  \expandafter\newcommand\csname #1check\endcsname{\textcolor{#1}{\ding{52}}}%
}

\newcommand*\xcheck[1]{%
  \expandafter\newcommand\csname #1check\endcsname{\textcolor{#1}{\ding{56}}}%
}

\colourcheck{green}
\xcheck{red}

\def\BibTeX{{\rm B\kern-.05em{\sc i\kern-.025em b}\kern-.08em
    T\kern-.1667em\lower.7ex\hbox{E}\kern-.125emX}}

\SetKwInput{KwInput}{Input}                % Set the Input
\SetKwInput{KwOutput}{Output}              % set the Output
\SetKwComment{Comment}{$\triangleright$\ }  {}

\begin{document}

\title{Federated Radio Frequency Fingerprinting with Model Transfer and Adaptation
}
\author[$\dag$]{Chuanting Zhang}
\author[$\dag$]{Shuping Dang}
\author[$\ddag$]{Junqing Zhang}
\author[$\S$]{Haixia Zhang}
\author[$\dag$]{Mark~A.~Beach}
\affil[$\dag$]{Department of Electrical and Electronic Engineering, University of Bristol, United Kingdom}
\affil[$\ddag$]{Department of Electrical Engineering and Electronics, University of Liverpool, United Kingdom}
\affil[$\S$]{School of Control Science and Engineering, Shandong University, Jinan, China}
% \affil[ ]{Email: chuanting.zhang@bristol.ac.uk}

\maketitle
\thispagestyle{plain}
\pagestyle{plain}

\begin{abstract}
The Radio frequency (RF) fingerprinting technique makes highly secure device authentication possible for future networks by exploiting hardware imperfections introduced during manufacturing. Although this technique has received considerable attention over the past few years, RF fingerprinting still faces great challenges of channel-variation-induced data distribution drifts between the training phase and the test phase. To address this fundamental challenge and support model training and testing at the edge, we propose a federated RF fingerprinting algorithm with a novel strategy called model transfer and adaptation (MTA). The proposed algorithm introduces dense connectivity among convolutional layers into RF fingerprinting to enhance learning accuracy and reduce model complexity. Besides, we implement the proposed algorithm in the context of federated learning, making our algorithm communication efficient and privacy-preserved. To further conquer the data mismatch challenge, we transfer the learned model from one channel condition and adapt it to other channel conditions with only a limited amount of information, leading to highly accurate predictions under environmental drifts. 
Experimental results on real-world datasets demonstrate that the proposed algorithm is model-agnostic and also signal-irrelevant. Compared with state-of-the-art RF fingerprinting algorithms, our algorithm can improve prediction performance considerably with a performance gain of up to 15\%.

\end{abstract}


\section{Introduction}
% Wireless networks are transforming almost every aspect of our life, from various entertainment activities we enjoy to the different ways we access daily news.
Wireless communications have been connecting massive Internet of things (IoT) devices and transforming our everyday life with numerous exciting applications including smart homes, smart cities, connected healthcare, etc. However, due to the broadcast nature of wireless transmissions, IoT devices always suffer from security vulnerabilities and face many attack vectors, such as phishing and spoofing \cite{zhang_magzine_2022}.
Device identification is crucial for protecting wireless networks since access requests can be denied if malicious devices or users are identified.
Traditional device identification schemes are mainly based on techniques stemming from cryptography, such as encryption using public or private keys to protect higher-layer information security and to achieve identity recognition. On the other hand, these techniques could fail if the key is compromised. Thus, new types of security mechanisms in the lower layer are needed to enhance wireless network security. 

Radio frequency (RF) fingerprinting is an emerging physical layer security technique that helps with device identification by exploiting hardware impairments that are  hidden in the electromagnetic waves of the transmitter \cite{Nair2022imc, Soltanieh2020rfi, Xing2018CL}. Recent research has found that every transmitter has its unique RF fingerprint resulting from imperfections of analog components, which are non-reproducible by attackers \cite{REHMAN2014591, Zou_jproc2016}.

% \begin{figure}[!t]
% \centering
% \begin{subfigure}{0.2\textwidth}
%     \includegraphics[width=\textwidth]{figs/distribution_mismatch_density_v2.pdf}
%     \caption{Data distribution mismatch.}
%     \label{fig:mismatch}
% \end{subfigure}
% \hfill
% \begin{subfigure}{0.2\textwidth}
%     \includegraphics[width=\textwidth]{figs/performance_drop_v2.pdf}
%     \caption{Performance degradation.}
%     \label{fig:drop}
% \end{subfigure}
% \caption{Illustration of data distribution mismatch between different days and its impacts on prediction performance: Training a model using data from day 1 and testing it on day 1 and day 2 yield distinct prediction performance.}
% \label{fig:motivation}
% \end{figure}

Implementing RF fingerprinting using machine learning algorithms has been the mainstream due to their effective non-linear relationship modeling between input features and device identity. 
The earlier works on RF fingerprinting focus mainly on using handcrafted features to carry out classification, relying heavily on domain knowledge and human labor to extract features \cite{Kennedy_vtc2008}. What is even worse, the significance of these features is non-deterministic and task-dependent. This dilemma can be greatly eased thanks to the advancement of deep learning techniques,
% explanation for Junqing: there are deep learning models that are not based on neural networks such as deep Boltzmann machines.
in particular, deep neural networks, making automatic feature learning possible. 
The work in~\cite{Riyaz2018} adopted a convolutional neural network (CNN) framework to identify the type of protocol in use and the specific radio transmitter. After this, research works based upon CNN for RF fingerprinting have flourished \cite{DeepRadioID2019,Guillem2020_infocom,shen_jsac_2021,Zhang_tifs_2021,Shen_tifs_2022,Soltani2020}. In \cite{DeepRadioID2019}, the authors proposed DeepRadioID, another CNN framework for device identification that is robust against different channel conditions with the help of a carefully optimized digital finite input response filter.
Similarly, to overcome the performance degradation caused by the randomness of wireless channels, other solutions, such as metric learning \cite{Guillem2020_infocom}, carrier frequency offset compensation \cite{shen_jsac_2021}, and short-time Fourier transformation \cite{Shen_tifs_2022}, were also explored in the existing literature. Specifically, these solutions were verified for different application scenarios, including unmanned aerial vehicle communication networks \cite{Soltani2020} and LoRa networks \cite{Zhang_tifs_2021}. Experimental results on large-scale datasets demonstrate that considerable performance gains can be harvested by using the CNN architecture compared with traditional learning algorithms \cite{Jian2020,Sankhe2020}.

Though many studies exist nowadays, RF fingerprinting still faces several challenges hindering its practical implementation. First, most of the works mentioned above focus on centralized learning and require data to be transferred to a server before processing, which is obviously inappropriate when data privacy and protection are mandatory \cite{Piva2021mobihoc}. 
Second, one of the fundamental challenges of RF fingerprinting is the data distribution drift between the training and testing phases, which is generally caused by environments. The drift could lead to a significant mismatch between training and testing data and degrade prediction performance considerably. How to solve this challenge is still yet to be resolved.
Last but not least,  training a deep CNN is time-consuming due to the requirement of adapting a huge number of network parameters. Designing an effective and efficient CNN architecture specifically for RF fingerprinting is also challenging.

This paper tries to address the above challenges and proposes a federated RF fingerprinting algorithm based on a strategy termed model transfer and adaptation (MTA). We design a CNN framework based on the dense connectivity \cite{zhang2018citywide,huang2017densely} with careful calibration, aiming to reduce model complexity. Besides, with MTA, our model can quickly adapt to the new data distribution using a limited number of data samples and requires no complex pre-processing of data. Our main contributions are summarized as follows:
\begin{itemize}
    \item We propose an RF fingerprint algorithm based upon the federated learning paradigm\cite{zhang2021fedda, pmlr-v54-mcmahan17a}, which does not require raw data collected in a centralized manner, thus guaranteeing data privacy.
    \item We introduce dense connectivity into RF fingerprinting and design a novel CNN framework that is capable of considerably reducing model complexity.
    \item To overcome the data distribution drift and the data mismatch problem, we propose MTA, a strategy that can fully leverage the knowledge of a learned model from one channel condition and quickly adapt it to new channel conditions requiring only a limited number of data samples. MTA is model-agnostic and can be applied to most learning algorithms.
    \item We verify our algorithm on real-world datasets, and the results demonstrate that our algorithm is superior to state-of-the-art RF fingerprinting methods.
\end{itemize}

\section{System Model and Problem Formulation}
In this section, we first give our considered system model and then formulate the research problem in the context of federated learning.

\subsection{System Model}
\begin{figure}[!t]
\centerline{\includegraphics[draft=false,width=0.6\textwidth]{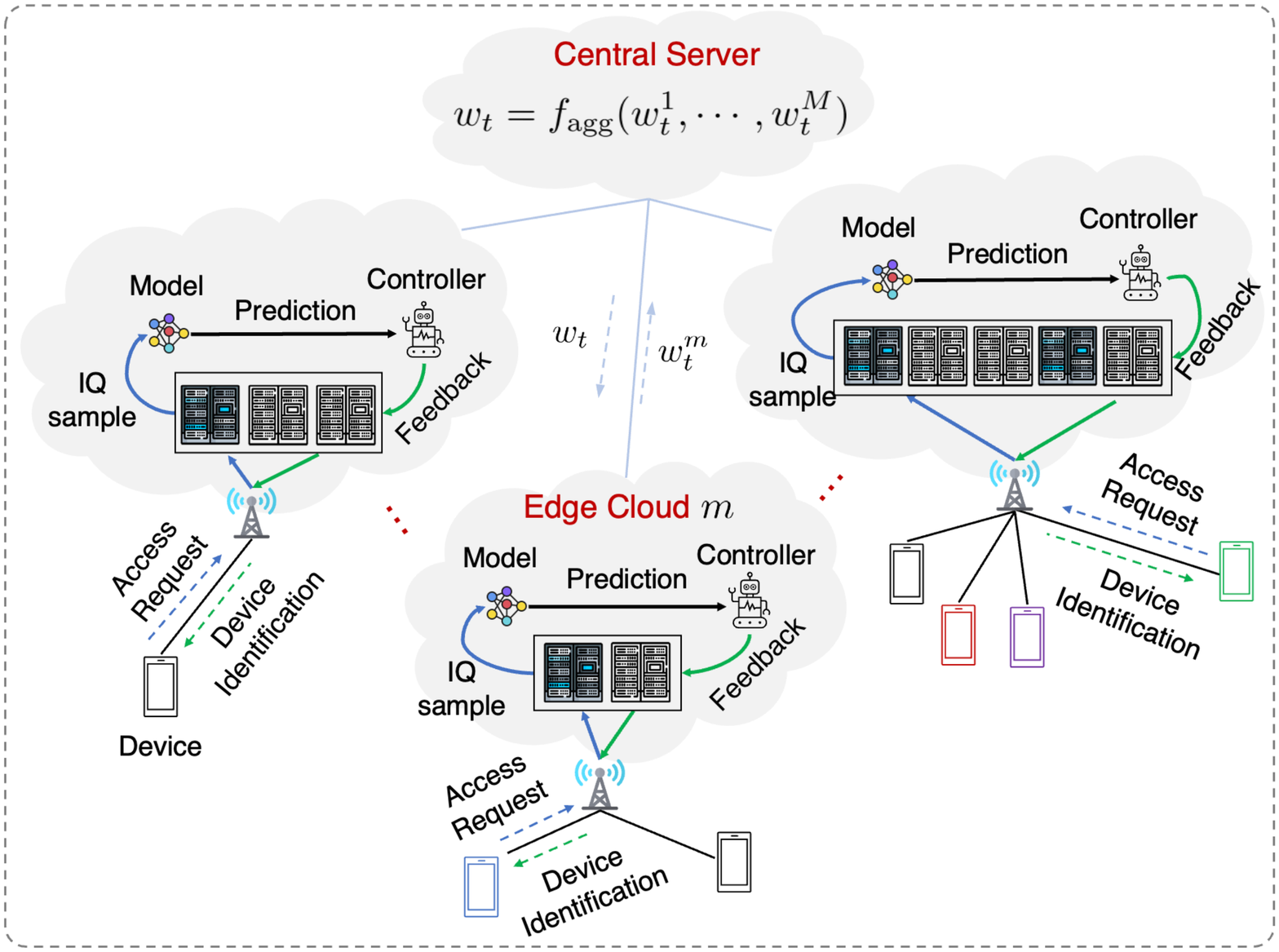}}
\caption{System model of the proposed federated RF fingerprinting.}
\label{fig:sys}
\end{figure}

In this paper, we consider the RF fingerprinting problem in future edge networks, as shown in Fig.~\ref{fig:sys}. There are $M$ separate edge cloud units that connect to a trusted central server. Each edge cloud unit possesses both communication and computing abilities and serves a group of user devices. With the consideration of privacy preservation, raw data transferring among edge cloud units is not permitted. Meanwhile, each edge unit $m\in\mathcal{M}=\{1,2,\dots,M\}$ is deployed with a prediction model which is parameterized with $w$. This model performs prediction whenever a user device makes access requests to the base station and judges if the device is from a legitimate user group or not. Generally speaking, each edge cloud unit $m$ has a local and private dataset denoted as $D_m=\{(\mathbf{x}_{m,i}, y_{m,i})\}_{i=1}^{N_m}$ with $\mathbf{x}_{m,i} \in \mathbb{R}^{C\times H}$ being features extracted from in-phase and quadrature (IQ) samples and $y_{m,i}$ the corresponding device identity. $C$ indicates the number of types of input adopted, and $H$ represents the dimension of the input. We detail the construction of $\mathbf{x}_{m,i}$ in Section \ref{tsc}.

\subsection{Problem Formulation}
Given a prediction model and each edge cloud unit's private dataset $D_m$, the prediction problem can be formulated as
\begin{equation}\label{obj}
    \arg\min_{w} \Big\{ f(w)  = \frac{1}{M} \sum_{m=1}^M f_m(w) \Big\},
\end{equation}
where $f_m(\cdot)$ denotes the local objective at the edge cloud unit $m$, which can be explicitly expressed as
\begin{equation}\label{local:obj}
       f_m(w) = \frac{1}{N_m} \sum_{i=1}^{N_m} \ell(\mathsf{d}_{m,i}; w),
\end{equation}
where $\mathsf{d}_{m,i} \in D_m$ denotes the input data samples selected from $D_m$, and $\ell(\cdot)$ is a customized loss function quantifying the accuracy of the prediction returned by the model and will be detailed in Section \ref{cnndc}.
In this paper, (\ref{obj}) is expected to be solved collaboratively by the $M$ edge cloud units as shown in Fig.~\ref{fig:sys}.

\begin{figure}[!t]
\centering
\begin{subfigure}{0.45\textwidth}
    \includegraphics[draft=false,width=\textwidth]{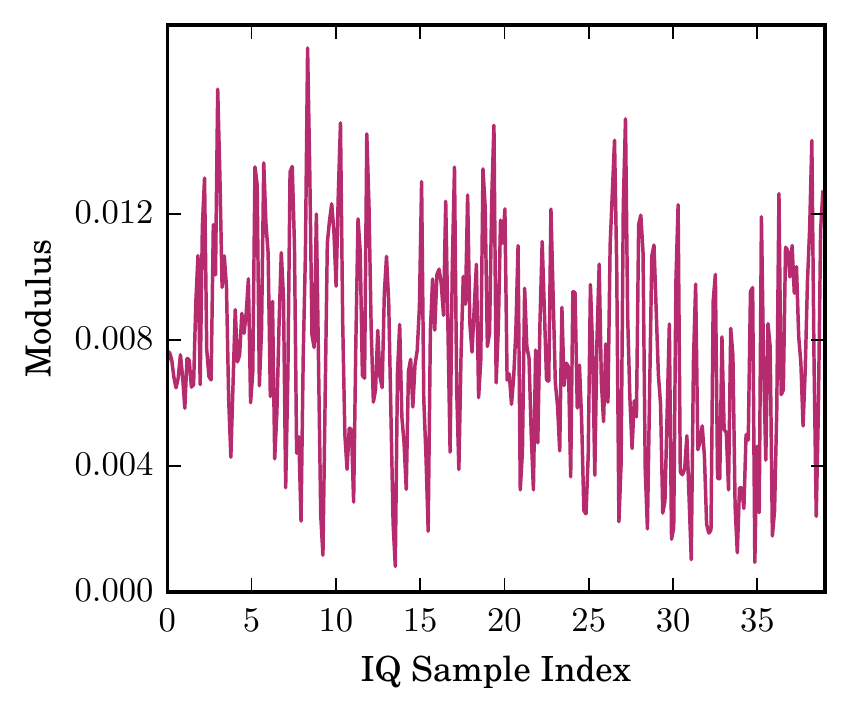}
    \caption{Temporal dynamics.}
    \label{fig:time}
\end{subfigure}
\begin{subfigure}{0.45\textwidth}
    \includegraphics[draft=false,width=\textwidth]{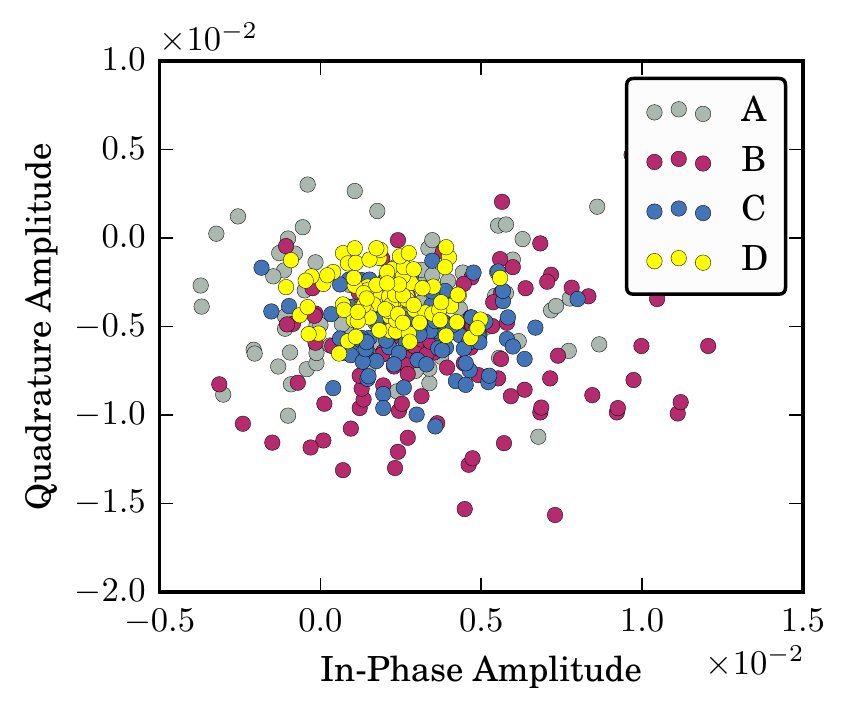}
    \caption{Spatial distribution.}
    \label{fig:iq}
\end{subfigure}
\caption{Visualization of IQ samples in the time and space domains.}
\label{fig:data}
\end{figure}

\section{CNN Framework Design for RF Fingerprinting}
% In this section, we first explain how to construct $\mathbf{x}_{m}$ and then propose the CNN framework design, i.e., a prediction model parameterized by $w$. After that, we present the MTA strategy to overcome the data distribution drift challenge confronted with RF fingerprinting and summarize the holistic process in a compact algorithm.

\subsection{Training Sample Construction}\label{tsc}
Given a stream of received IQ samples $\mathbf{s}$, whose value at time slot $t$ is given by $s_t=I_t + jQ_t$, where $I_t$ and $Q_t$ are two real values denoting the in-phase and quadrature amplitudes of the RF signal. 
A sample IQ stream is illustrated in Fig.~\ref{fig:data}. 
By using the sliding window scheme, we first generate a series of data segments with the window size $H$, then extract the in-phase branch ($I_t$), quadrature branch ($Q_t$), and the corresponding modulus branch ($\sqrt{I_t^2 + Q_t^2}$), and treat them as input features to our prediction model. Thus, the types of input features ($C$) are 3. In this paper, $H$ is set to $1024$ since we empirically found that larger values yield no performance improvements yet raise computational complexity.

\subsection{CNN Framework with Dense Connectivity}\label{cnndc}
\begin{figure}[!t]
\centerline{\includegraphics[draft=false,width=0.6\textwidth]{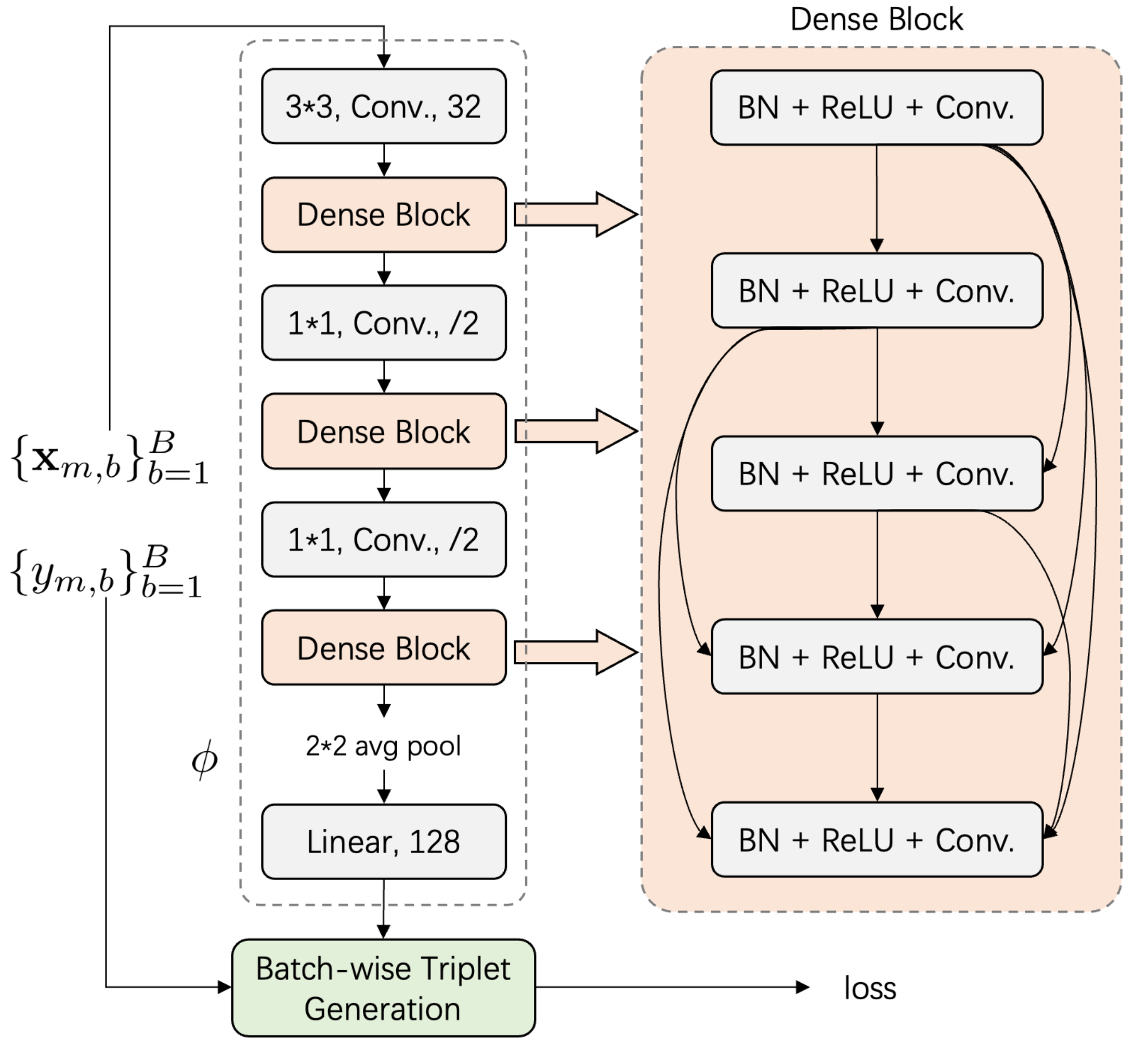}}
\caption{Proposed framework for RF fingerprinting.}
\label{fig:densenet}
\end{figure}

Compared with traditional RF fingerprinting systems that adopt residual connectivity for learning RF features, we resort to dense connectivity \cite{huang2017densely} for enhancing feature reuse and propagation between different layers. 
As shown in Fig.~\ref{fig:densenet}, our proposed CNN framework mainly consists of three convolutional layers, three dense blocks, and a linear layer, followed by a batch-wise triplet generation module. For convolutional layers, their kernel sizes vary according to functional differences and so does the corresponding output feature mapping relation.
Each dense block comprises $L$ layers with each layer $l$ being a composite function $\mathcal{F}_l(\cdot)$ that sequentially performs operations of batch normalization (BN), ReLU activation, and convolution. For the $l$-th layer of a dense block, the output feature mapping can be expressed as
\begin{equation}
    \mathbf{x}_{l} = \mathcal{F}_l(\mathbf{x}_0 \oplus \mathbf{x}_1 \oplus \cdots \oplus \mathbf{x}_{l-1}),
\end{equation}
where $\mathbf{x}_0$ denotes the initial input to a dense block, and $\oplus$ is the concatenation operation.
It should be noted that concatenation operation can greatly facilitate feature propagation and reuse between different layers, mitigating the gradient vanishing and exploding problems as an additional benefit.

Since a dense block has many layers and each layer will transfer its feature mapping relation to all its subsequent layers, the model complexity increases exponentially and the training process could be quite time-consuming. To mitigate this drawback, we set the kernel size of the convolutional layer between dense blocks to $1\times 1$ and also reduce the feature mapping relation to half of its input. We denote the framework as a function $\phi$, then for batch input $\mathbf{x}_{m,b} \in D_m$, the output feature vectors from the final linear layer can be written as $\{\phi(\mathbf{x}_{m,b})\}_{b=1}^B$. 
% \jz{$\mathbf{x}_{m,b}$ is not defined yet.}

\begin{algorithm}[!t]
\KwInput{IQ samples $\{D_m\}_{m=1}^M$;~training round $R$; ~learning rate $\eta$; initialized global model $w_0$}
\KwOutput{Global model: $w$}
\For {$r=0,1, \cdots, R$ }
		{
			Select a subset of edge cloud units $\mathcal{M}_r$ \\
% 			$\mathsf{M} \leftarrow max (M\cdot \delta, 1)$ \\
% 			$M_t \leftarrow$ a random set of $\mathsf{M}$ edge cloud unit\\
            \Comment {run on edge cloud units in parallel}
			\For {$m \in \mathcal{M}_r$}
			{
			    Receive global model and $w_{r}^m \leftarrow w_{r}$; \\
			    \For {each local epoch $k=0,1,\cdots$}
			    {
			        \For{each batch $\mathcal{B}$ in $D_m$}
			        {
			            $w_{r}^m \leftarrow w_{r}^m - \eta \nabla f(\mathcal{B};w_{r}^m)$;
			        }
			    }
			    $w_{r+1}^m \leftarrow w_r^m$ and upload it to the central server; \\
			}
			\Comment{run on server side}
			Local model aggregation: $w_{r+1} = \frac{1}{|\mathcal{M}_r|} \sum_{m\in \mathcal{M}_r} w_{r+1}^m$;
		}

\caption{Federated RF Fingerprinting}
\label{alg:fedrf}
\end{algorithm}

% \subsection{Triplet Generation and Loss Function}\label{tglf}
After obtaining the feature vectors, we construct $B$ triplets $\{\mathbf{a}_{m,b}, \mathbf{p}_{m,b}, \mathbf{n}_{m,b}\}_{b=1}^B$ based on the label information $\{y_{m,b}\}_{b=1}^B$. Specifically, $\mathbf{a}_{m,b}$ is associated with $\mathsf{Anchor}$, and its value is exactly $\phi(\mathbf{x}_{m,b})$; $\mathbf{p}_{m,b}$ is relevant to $\mathsf{Positive}$ which is sampled from $\{\phi(\mathbf{x}_{m,b})\}_{b=1}^B$, and its corresponding label is the same as $\mathbf{a}_{m,b}$. Similarly, $\mathbf{n}_{m,b}$ pertains to $\mathsf{Negative}$, which is also sampled from $\{\phi(\mathbf{x}_{m,b})\}_{b=1}^B$, and its corresponding label is different from $\mathbf{a}_{m,b}$. With the defined triplets, the loss function is computed by
\begin{equation}
    \ell (\mathbf{a}_{m,b}, \mathbf{p}_{m,b}, \mathbf{n}_{m,b}) = \text{max}\{d_{ap} - d_{an} + \alpha, 0\},
\end{equation}
where $\alpha$ is a preset threshold that controls the flexibility; $d_{ap}$ and $d_{an}$ represent the $\mathsf{Anchor}$-$\mathsf{Positive}$ and $\mathsf{Anchor}$-$\mathsf{Negative}$ distances, respectively, which are defined as
\begin{equation}
\begin{split}
    d_{ap} = || \mathbf{a}_{m,b}- \mathbf{p}_{m,b}||_2^2, \\
    d_{an} = || \mathbf{a}_{m,b}- \mathbf{n}_{m,b}||_2^2.
\end{split}
\end{equation}
Local model parameter $w$ can be obtained by optimizing the above loss function.

\subsection{Federated Training}
Having data prepared and the RF fingerprinting framework constructed, we can proceed with the model training. The training procedure is summarized in Algorithm~\ref{alg:fedrf} and generally consists of three steps:
\begin{itemize}
    \item A subset edge cloud units $\mathcal{M}_t$ is selected and the central server broadcasts the global model $w_r$ to each edge cloud unit of $\mathcal{M}_t$ at the beginning of the $r$-th round training;
    \item Based upon the received global model, each edge cloud unit updates $w_r$ using its local and private data. The one-step update rule can be written as 
    \begin{equation}
        w_{r+1}^m \leftarrow w_r - \eta \nabla f(\mathcal{B}^m_r;w_r),
    \end{equation}
    where $\eta$ is the local learning rate, and $\mathcal{B}^m_r$ is the batch input data. The local model will be uploaded to the central server once the updating process is complete.
    % explanation for Junqing: there are many kinds of parameter aggregation and Averaging is one of them.
    \item The central server aggregates local models to yield a new global model by the following relation:
    \begin{equation}
        w_{r+1} = \frac{1}{|\mathcal{M}_r|} \sum_{m\in \mathcal{M}_r} w_{r+1}^m.
    \end{equation}
\end{itemize}

These three steps will be iterated until the loss function converges or the training process reaches the maximum round. After that, we obtain the final model parameter $w$ and use it to make the prediction.

\subsection{Model Transfer and Adaptation}
In the last subsection, we explained how we obtain model global $w$. When deploying the model in practical environments to conduct production, its performance will be significantly challenged when the environment drifts since the data samples from the new environment could be rather different from those used to train the model. To alleviate this problem, we propose a strategy called MTA based on the idea of transfer learning \cite{zhang2019deep}. The MTA strategy works as follows: when the system encounters a different environment with new data, we do not retrain the model from scratch. Instead, we transfer the learned model $w$ to this new environment and use it as the initialized global model. Consequently, we can benefit from this well-designed initialization and easily adapt the global model using a limited number of IQ samples for only several rounds, producing an updated global model $w'$. Then we use $w'$ to identify the labels of new IQ samples.

\subsection{Model Prediction}
Once new data $\mathbf{x}$ comes, we input it into the model and obtain its feature $\phi(\mathbf{x})$. The distances between $\phi(\mathbf{x})$ and each class's feature vector will be calculated and compared. The label of $\phi(\mathbf{x})$ will be predicted as the one with the smallest distance from it.

\section{Experimental Evaluation}
% This section devotes to the explanation of experimental settings and obtained results. The parameters related to federated learning and the CNN framework are first given. Then we report detailed prediction performance comparisons between our proposed algorithm and baselines. 
% Particularly, we present the effectiveness of our MTA strategy when being applied to different learning algorithms.
\subsection{Setup}
\subsubsection{Dataset and Parameter Settings}
We test the performance of our algorithm on a public benchmark dataset released in \cite{Guillem2020_infocom}. The dataset includes four devices' three different kinds of IQ samples pertaining to WiFi, 4G LTE, and 5G NR. The four devices (denoted as A, B, C, and D) are USRP X310 transmitters and the receiver is a USRP B210 platform. The signal samples are random payloads simulated using MATLAB toolboxes and are logged on two different days with different channel environments. A visualization of the IQ samples has been shown in Fig.~\ref{fig:data}, from which we can notice that different devices' IQ samples are coupled with each other, and simple linear models could not effectively distinguish them. In our simulations, the first day's IQ samples will be used as the training dataset. For the second day's IQ samples, we select the first $\rho$ samples to perform the MTA strategy and test the performance on the rest of the IQ samples.

Both the training and test datasets have $12000$ samples. All data is distributed into $100$ edge cloud units to imitate a federated learning scenario. The model is trained locally using the Adam optimizer for $10$ epochs with a batch size $10$. We set the learning rate to $0.0001$ and the global round to $50$. In each global round, we suppose 10\% of edge cloud units are involved in training due to different traffic loads.

\subsubsection{Proposed and Baseline Algorithms}
For our algorithm, each dense block has $6$ convolutional layers. There are two versions of our algorithm sharing the same network architecture, i.e., Proposed-Basic (w/o MTA strategy) and Proposed-MTA (w/ MTA strategy). We compare our algorithms with the following baselines.

\begin{itemize}
    \item Federated MLP: An architecture that has two linear layers with hidden sizes of $512$ and $4$.
	\item Federated CNN. An architecture that has two convolutional layers and two linear layers. The hidden sizes are $16$, $64$, $512$, and $4$, respectively.
	\item Federated ResNet. An architecture that has four convolutional blocks and one linear layer. Each block consists of four convolutional layers with feature maps of $32$, and the last linear layer has a hidden size of $4$.
	\item Centralized ResNet. This algorithm has exactly the same architecture as the above ResNet but is trained in a centralized way.
\end{itemize}
Unless otherwise stated, the kernel sizes of hidden layers and the activation functions in both our algorithms and the baselines are $3\times 3$ and ReLU, respectively.

\subsection{Overall Prediction Performance}
Table~\ref{tab:table1} presents a detailed performance comparison between our proposed algorithms and those baselines.
The prediction performance is measured in terms of accuracy, and a greater value indicates better prediction.  
We can observe from Table~\ref{tab:table1} that our RF fingerprinting algorithm achieves the highest prediction accuracy on all datasets of different signal samples, compared with all baselines regardless of centralized training or federated training. There is a considerable performance improvement in our results over the second-best ones. For example, the obtained accuracy of $0.9343$ on the 4G signal and $0.9033$ on the hybrid signal (a mix of 4G, 5G, and WiFi signals) samples are much higher than the accuracy of $0.8257$ and $0.7692$ pertaining to the centralized ResNet and the federated ResNet, respectively. Even without using the MTA strategy, our designed CNN framework with dense connectivity still achieves better predictions than baselines. The results indicate that approximately $10\%$ to $15\%$ improvements can be achieved by our proposed algorithm.
A more noticeable observation is that our algorithm has roughly $80,000$ parameters, which are less than those of the baselines. In a nutshell, the results presented in Table~\ref{tab:table1} demonstrate the effectiveness of dense connectivity and the MTA strategy for RF fingerprinting.
\begin{table*}[!t]
\centering
\renewcommand{\arraystretch}{1.2}
\caption{Performance comparison between the proposed algorithm and baselines.\label{tab:table1}}
\begin{tabular}{|c|c|c|c|c|c|c|c|}
\hline
                         &                      & Centralized ResNet   & Federated MLP    & Federated CNN    & Federated ResNet & Proposed-Basic & Proposed-MTA   \\ \hline
\multirow{2}{*}{Notes}   & $\#$ of Param. & $157.54$ K & $1.58$ M    & $1.19$ M &  $157.54$ K  & \multicolumn{2}{c|}{$79.95$ K}   \\ \cline{2-8}
                         & Federated            &  \redcheck    &\greencheck &\greencheck&    \greencheck  &  \multicolumn{2}{c|}{ \greencheck}   \\ \hline
\multirow{4}{*}{Signals} & 4G                   & $0.8257$   & $0.7110$ & $0.7895$  & $0.7245$  & \underline{$0.8683$} & $\textbf{0.9343}$  \\
                         & 5G                   & $0.8375$   & $0.7123$ & $0.7725$  & $0.7448$  & $\textbf{0.9105}$ & $\underline{0.9100}$  \\
                         & WiFi                 & \underline{$0.9690$}   & $0.7205$ & $0.8213$  & $0.7688$  & $0.7508$ & $\textbf{0.9800}$ \\
                         & Hybrid               & $0.7268$   & $0.6923$ & $0.7553$  & $0.7692$  & \underline{$0.7721$} & $\textbf{0.9003}$  \\ \hline
\end{tabular}
\end{table*}

\subsection{Per-Device Prediction Performance}
The previous subsection reports the overall prediction performance for different types of signal samples while this section presents the prediction results for each device. We only compare our proposed algorithm with federated ResNet on the 4G and hybrid signal datasets for simplicity. 

\begin{figure*}[!t]
\centering
\begin{subfigure}{0.3\textwidth}
    \includegraphics[draft=false,width=\textwidth]{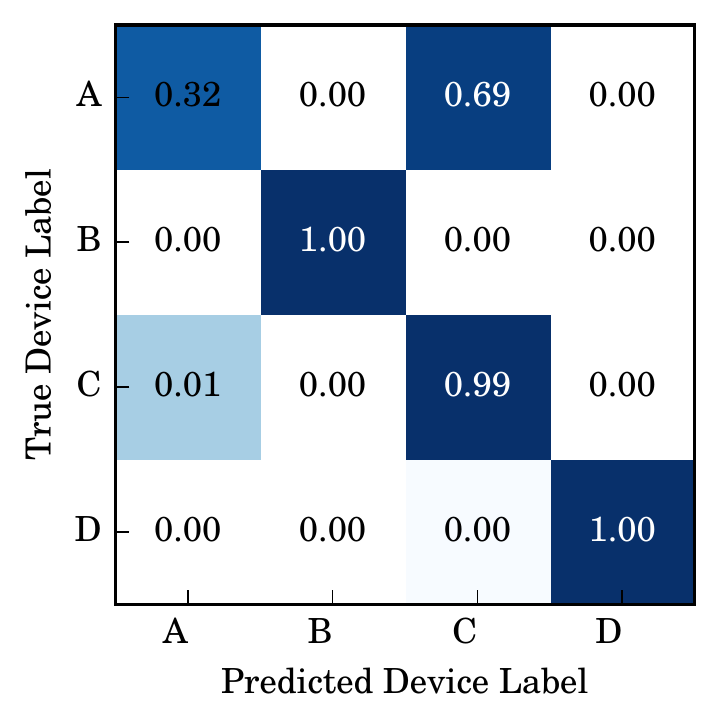}
    \caption{Centralized ResNet (4G).}
    \label{fig:res_cm_4g}
\end{subfigure}
\hspace{5mm}
\begin{subfigure}{0.3\textwidth}
    \includegraphics[draft=false,width=\textwidth]{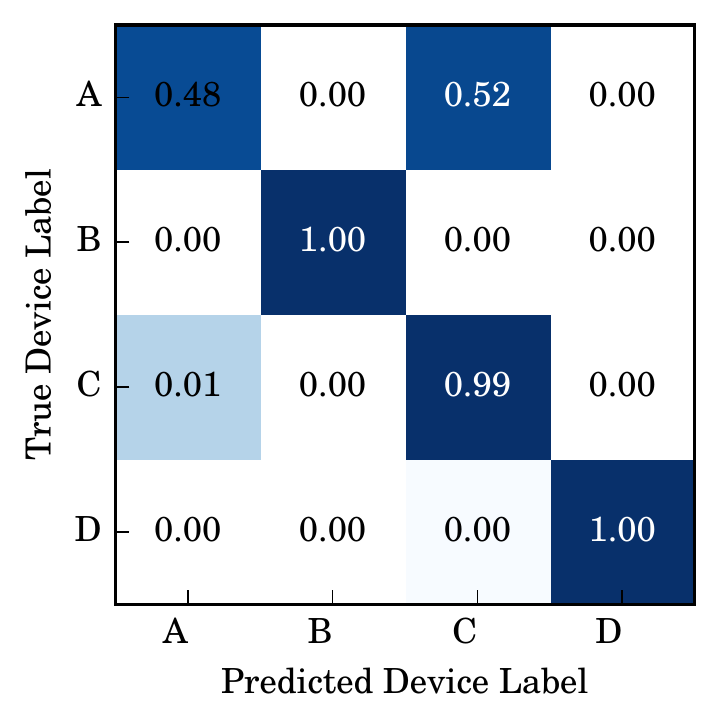}
    \caption{Proposed-Basic (4G).}
    \label{fig:basic_cm_4g}
\end{subfigure}
\hspace{5mm}
\begin{subfigure}{0.3\textwidth}
    \includegraphics[draft=false,width=\textwidth]{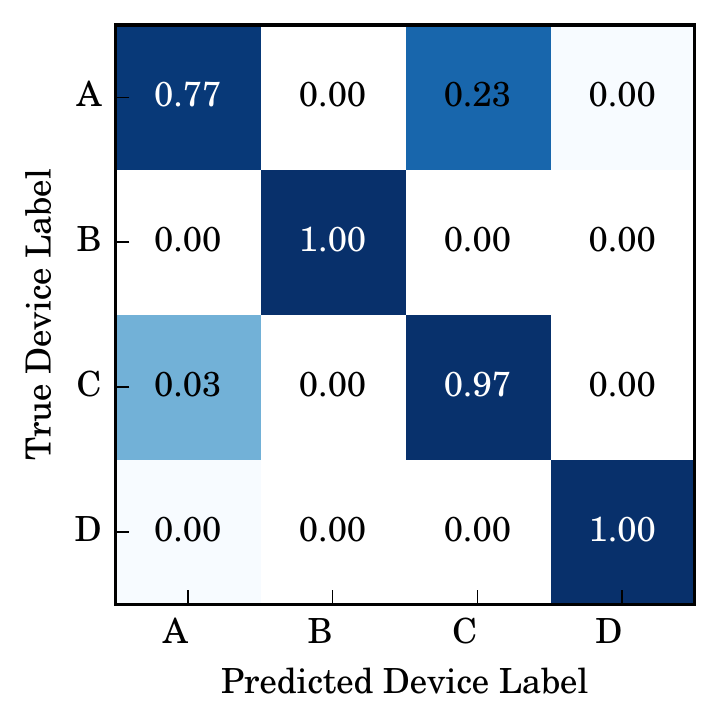}
    \caption{Proposed-MTA (4G).}
    \label{fig:mta_cm_4g}
\end{subfigure}

\begin{subfigure}{0.3\textwidth}
    \includegraphics[draft=false,width=\textwidth]{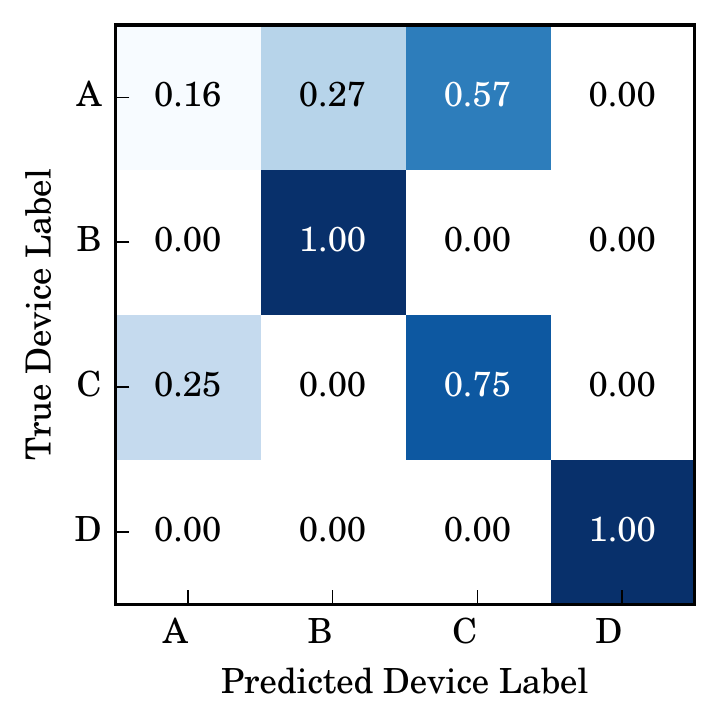}
    \caption{Centralized ResNet (Hybrid).}
    \label{fig:res_cm_all}
\end{subfigure}
\hspace{5mm}
\begin{subfigure}{0.3\textwidth}
    \includegraphics[draft=false,width=\textwidth]{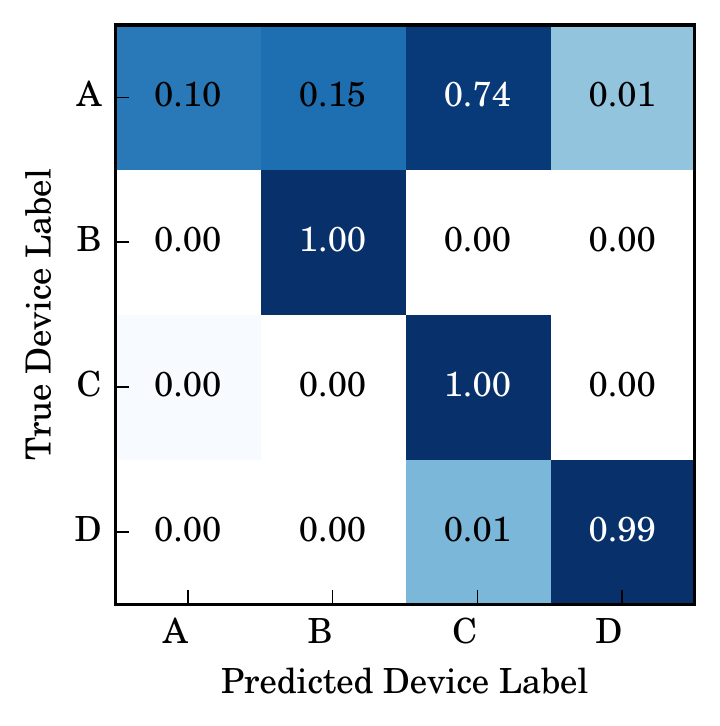}
    \caption{Proposed-Basic (Hybrid).}
    \label{fig:basic_cm_all}
\end{subfigure}
\hspace{5mm}
\begin{subfigure}{0.3\textwidth}
    \includegraphics[draft=false,width=\textwidth]{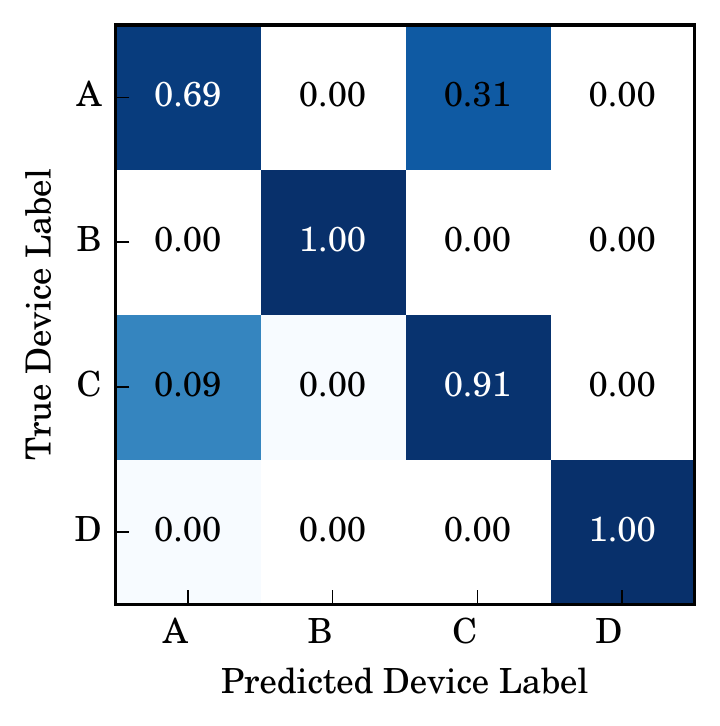}
    \caption{Proposed-MTA (Hybrid).}
    \label{fig:mta_cm_all}
\end{subfigure}

\caption{Confusion matrix of centralized ResNet and the proposed algorithms for various signal datasets.}
\label{fig:confusion_matrix}
\end{figure*}

The obtained results are displayed in Fig.~\ref{fig:confusion_matrix}. Both federated ResNet and the proposed algorithm perform well for devices B, C, and D on the 4G signal samples, achieving almost $100\%$ accuracy. But federated ResNet's accuracy for device A is only $32\%$, which may be because device A is quite similar to device C.
However, the obtained results of the proposed algorithm for device A are $48\%$ (w/o MTA) and $77\%$ (w/ MTA). For the hybrid signal dataset, a similar phenomenon can be observed, and we omit the repeated explanations here for simplicity.
Through in-depth analysis and comparison of Fig.~\ref{fig:confusion_matrix}, we can substantiate that our proposed algorithm, in particular with the MTA strategy, consistently outperforms the baseline algorithms in terms of per-device prediction. 

\subsection{Generalization Ability of MTA}
\begin{figure}[!t]
\centering
\begin{subfigure}{0.45\textwidth}
    \includegraphics[draft=false,width=\textwidth]{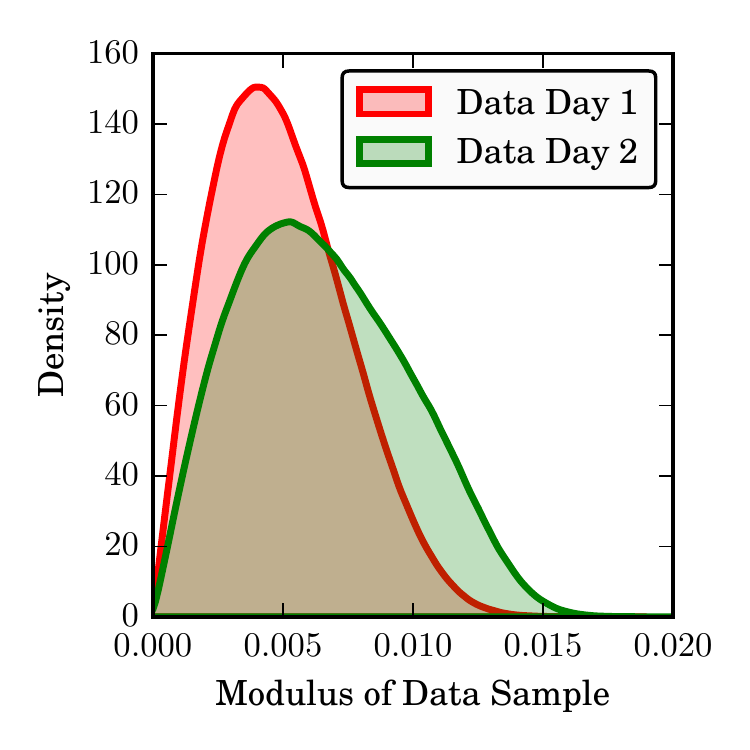}
    \caption{Data distribution mismatch.}
    \label{fig:mismatch}
\end{subfigure}
\begin{subfigure}{0.45\textwidth}
    \includegraphics[draft=false,width=\textwidth]{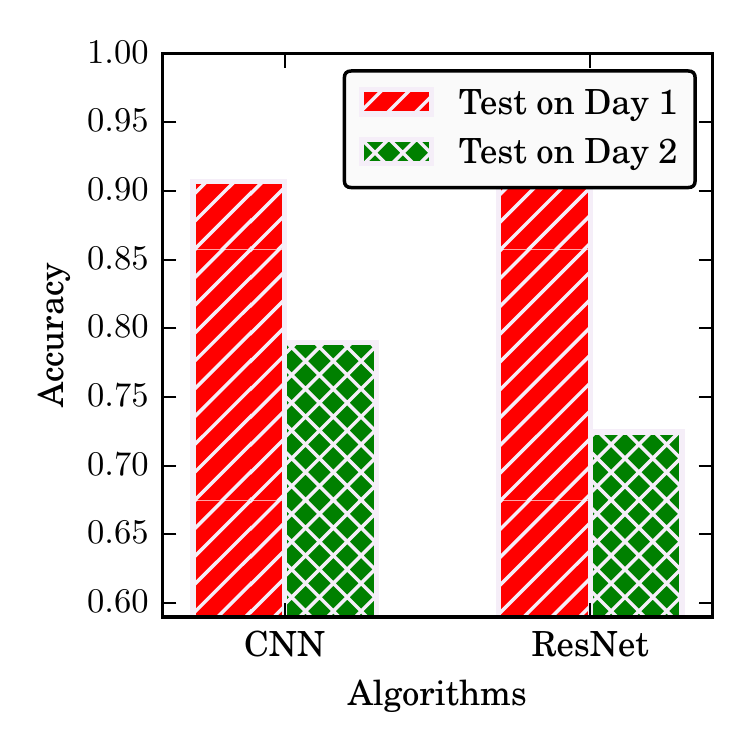}
    \caption{Performance degradation.}
    \label{fig:drop}
\end{subfigure}
\caption{Illustration of data distribution mismatch between different days and its impacts on prediction performance.
% : Training a model using data from day 1 and testing it on day 1 and day 2 yield distinct prediction performance.
}
\label{fig:motivation}
\end{figure}

\begin{figure}[!t]
\centering
\begin{subfigure}{0.45\textwidth}
    \includegraphics[draft=false,width=\textwidth]{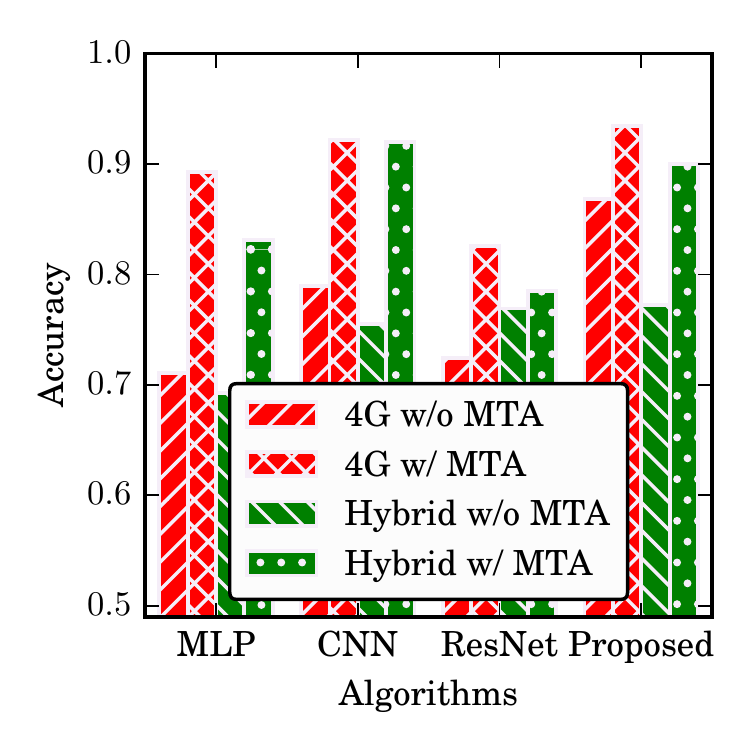}
    \caption{Predictions w/ and w/o MTA.}
    \label{fig:mta}
\end{subfigure}
% \hspace{5mm}
\begin{subfigure}{0.45\textwidth}
    \includegraphics[draft=false,width=\textwidth]{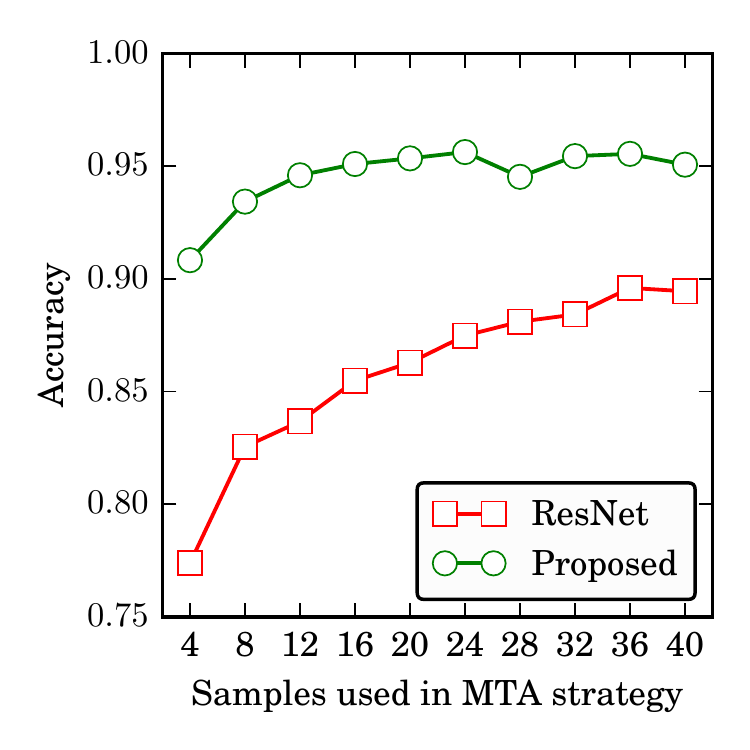}
    \caption{Predictions versus $\rho$.}
    \label{fig:alpha}
\end{subfigure}
\caption{Generalization ability of MTA.}
\label{fig:generalization}
\end{figure}

Training a model on one channel condition and testing it on another will degrade the prediction performance considerably due to the data distribution mismatch between different environments, as shown in Fig. \ref{fig:motivation}. The training accuracies of federated CNN and ResNet are up to $0.9$, but their test accuracies are lower than $0.8$. This phenomenon inspired us to propose the MTA strategy.
To demonstrate the generalization ability of our MTA strategy, we apply it to all algorithms and compare the prediction performance with and without MTA in Fig.~\ref{fig:mta}. It is evident that the MTA strategy can consistently boost the corresponding algorithm's performance. Taking the Federated ResNet algorithm as an example, its accuracy with respect to the 4G signal dataset and the hybrid signal dataset is improved from $0.7245$ and $0.7692$ to $0.8255$ and $0.7847$, respectively. In addition, Fig.~\ref{fig:alpha} displays the prediction performance of federated ResNet and our proposed algorithm when increasing the number of data samples used in the MTA strategy. We can observe that the accuracy gradually increases when more data samples are available for model adaptation.
%Thus, from the results of Fig.~\ref{fig:generalization}, we can verify that our MTA strategy is model-agnostic and can be seamlessly integrated with any pre-trained models, leading to considerable performance improvements for RF fingerprinting.

% \begin{figure}[!t]
% \centerline{\includegraphics[width=0.25\textwidth]{figs/fig_mta_v2.pdf}}
% \caption{Performance of the MTA strategy.}
% \label{fig:mta}
% \end{figure}

% \begin{figure}[!t]
% \centerline{\includegraphics[width=0.25\textwidth]{figs/alpha_v2.pdf}}
% \caption{Prediction performance versus the number of samples used by the MTA strategy.}
% \label{fig:alpha}
% \end{figure}

\section{Conclusion}
This paper investigated the RF fingerprinting problem in the context of federated learning for edge networks. We proposed a novel CNN framework by introducing dense connectivity into RF fingerprinting. We found that because of time-varying wireless environments, there could exist data distribution drifts between training and testing datasets, further leading to data distribution mismatches and degraded prediction performance. Accordingly, we designed an MTA strategy to mitigate this challenge and verified that the prediction performance can be considerably improved. 
More importantly, our federated training involved no data transferring among different edge cloud units, thus the privacy and protection of IQ dataset were guaranteed.
Possible future research directions include further reducing the communication workload between the edge cloud and central server by introducing model compression and enhancing the security of model transferring by adopting differential privacy techniques.

\section*{Acknowledgment}
% This work was supported by the UKRI/EPSRC SWAN Prosperity Partnership (EP/T005572/1).
This research is funded through the UKRI/EPSRC Prosperity Partnership in Secure Wireless Agile Networks (SWAN) EP/T005572/1, and its official website can be accessed through https://www.swan-partnership.ac.uk/.

\bibliographystyle{IEEEtran}
\bibliography{IEEEabrv,lib}

\end{document}